\documentclass[10pt,twocolumn,letterpaper]{article}

\usepackage{cvpr}              

\usepackage[dvipsnames]{xcolor}

\definecolor{cvprblue}{rgb}{0.21,0.49,0.74}

\usepackage[pagebackref,breaklinks,colorlinks,citecolor=cvprblue]{hyperref}
\usepackage{amsmath, amssymb}
\usepackage{graphicx}
\usepackage{amsmath}
\usepackage{mathtools}
\usepackage{multirow}
\usepackage{colortbl}
\usepackage{wrapfig}
\usepackage{hhline}
\usepackage[utf8]{inputenc} 
\usepackage[T1]{fontenc}    
\usepackage{hyperref}       
\usepackage{url}            
\usepackage{booktabs}       
\usepackage{amsfonts}       
\usepackage{nicefrac}       
\usepackage{microtype}      
\usepackage{xcolor}         
\usepackage{makecell}
\usepackage{float}
\usepackage{graphicx}
\usepackage{svg}
\usepackage{amsmath}
\usepackage{algorithm}
\usepackage{algpseudocode}
\usepackage{adjustbox}

\usepackage{verbatim}
\usepackage{amsthm}

\usepackage[dvipsnames]{xcolor}
\definecolor{mydarkgreen}{rgb}{0.0, 0.5, 0.0} 
\definecolor{mydarkred}{rgb}{0.6, 0.0, 0.0} 
\newcommand{\better}[1]{\,\textcolor{mydarkgreen}{\scriptsize(+#1)}}

\newcommand{\mypar}[1]{\vspace{4pt}\noindent\textbf{#1}~}

\newcommand{\Real}{\mathbb{R}}

\newcommand{\AAA}{\mathbf{A}}
\newcommand{\BB}{\mathbf{B}}
\newcommand{\CC}{\mathbf{C}}
\newcommand{\KK}{\mathbf{K}}
\newcommand{\WW}{\mathbf{W}}
\newcommand{\DD}{\mathbf{D}}

\newcommand{\PP}{\mathbf{P}}
\newcommand{\LL}{\mathbf{L}}
\newcommand{\II}{\mathbf{I}}
\newcommand{\xx}{\mathbf{x}}
\newcommand{\ff}{\mathbf{f}}
\newcommand{\uu}{\mathbf{u}}
\newcommand{\vv}{\mathbf{v}}
\newcommand{\yy}{\mathbf{y}}
\newcommand{\hh}{\mathbf{h}}

\newcommand{\phz}{\phantom{0}}

\newtheorem{theorem}{Theorem}[section]

\newcommand{\changes}[1]{\textcolor{black}{#1}}
\usepackage{glossaries}

\newacronym{cnns}{CNNs}{Convolutional Neural Networks}
\newacronym{cnn}{CNN}{Convolutional Neural Network}
\newacronym{densenet}{DenseNet}{Dense Convolutional Network}
\newacronym{vits}{ViTs}{Vision Transformers}
\newacronym{vit}{ViT}{Vision Transformer}
\newacronym{nlp}{NLP}{Natural Language Processing}
\newacronym{ssm}{SSM}{State Space Model}
\newacronym{ssms}{SSMs}{State Space Models}
\newacronym{deit}{DeiT}{Data-efficient image Transformers}
\newacronym{ss2d}{SS2D}{2D Selective Scan}
\newacronym{dpct}{DPCT}{Dynamic Patch Connectivity Traversal}
\newacronym{csm}{CSM}{Corss-Scan Module}
\newacronym{iam}{IAM}{Isometric Adjustment Module}
\newacronym{ema}{EMA}{Exponential Moving Average}
\newacronym{knn}{KNN}{K-Nearest Neighbors}
\newacronym{miou}{mIoU}{mean Intersection over Union}
\newacronym{vim}{ViM}{Vision Mamba}
\newacronym{msv}{MSVMamba}{Multi-Scale VMamba}
\newacronym{zoh}{ZOH}{zero-order hold}
\newacronym{rfn}{RFN}{Rotational Feature Normalizer}
\newacronym{sts}{STS}{Spectral Traversal Scan}
\newacronym{stm}{STM}{Spectral Traversal Merge}
\newacronym{erf}{ERF}{Effective Receptive Field}

\def\paperID{13761} 
\def\confName{CVPR}
\def\confYear{2025}

\title{Spectral State Space Model for Rotation-Invariant~Visual~Representation~Learning}

\author{
Sahar Dastani$^{1,2}$\thanks{correspondence: \texttt{sahar.dastani-oghani.1@ens.etsmtl.ca}}  \and 
Ali Bahri$^{1}$ \and 
Moslem Yazdanpanah$^{1}$ \and
Mehrdad Noori$^{1}$ \and   
David Osowiechi$^{1}$ \and  
Gustavo Adolfo Vargas Hakim$^{1}$ \and 
Farzad Beizaee$^{1}$ \and 
Milad Cheraghalikhani$^{1}$ \and 
Arnab Kumar Mondal$^{3}$\thanks{work done during author's PhD at Mila/McGill} \and
Herve Lombaert$^{2,4}$ \and 
Christian Desrosiers$^{1}$\\
$^1$LIVIA, ILLS, ÉTS Montréal, Canada, $^2$Mila - Quebec AI Institute, \\ $^3$Apple, $^4$Polytechnique Montreal\\ 
}

\begin{document}
\maketitle

\begin{abstract}
State Space Models (SSMs) have recently emerged as an alternative to Vision Transformers (ViTs) due to their unique ability of modeling global relationships with linear complexity. SSMs are specifically designed to capture spatially proximate relationships of image patches. However, they fail to identify relationships between conceptually related yet not adjacent patches. This limitation arises from the non-causal nature of image data, which lacks inherent directional relationships. Additionally, current vision-based SSMs are highly sensitive to transformations such as rotation. Their predefined scanning directions depend on the original image orientation, which can cause the model to produce inconsistent patch-processing sequences after rotation. To address these limitations, we introduce Spectral VMamba, a novel approach that effectively captures the global structure within an image by leveraging spectral information derived from the graph Laplacian of image patches. Through spectral decomposition, our approach encodes patch relationships independently of image orientation, achieving \changes{patch traversal} rotation invariance with our Rotational Feature Normalizer (RFN) module. Our experiments on classification tasks show that Spectral VMamba outperforms the leading SSM models in vision, such as VMamba, while maintaining invariance to rotations and a providing a similar runtime efficiency. \changes{The implementation is available at: \url{https://github.com/Sahardastani/spectral_vmamba.git}.}

\end{abstract}    
\vspace{-20pt}
\section{Introduction}
\label{sec:intro}

\begin{figure}[t]
    \centering
    \includegraphics[width=.98\linewidth]{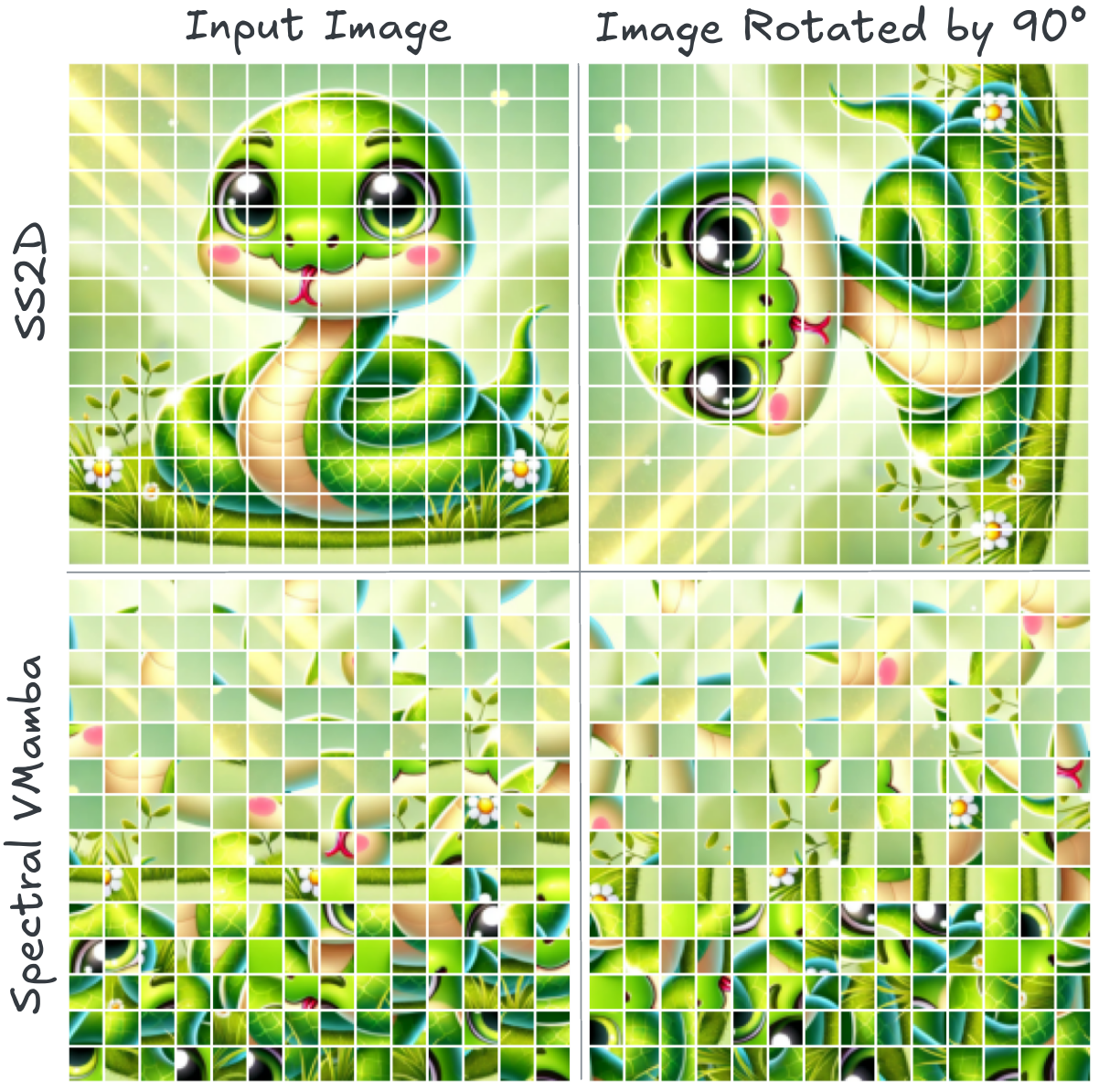}
    \caption{Effect of image rotation on patch processing in VMamba and Spectral VMamba networks. In VMamba networks (first row), patches are traversed using predefined horizontal and vertical scanning routes. As a result, rotating the image by 90° significantly changes the sequence in which patches are processed. In contrast, Spectral VMamba (second row) organizes patches using spectral information derived from a graph laplacian of the image patches. As shown in the second row, Spectral VMamba consistently processes background patches before snake patches in both the original and rotated images. The snake image in this analysis is AI-generated.}
    \label{patches}
\end{figure}

Visual representation learning is fundamental to computer vision, where precise feature extraction is essential for tasks such as image classification, detection, and segmentation. The ability to extract meaningful patterns directly impacts performance and unlocks deeper insights into visual tasks. Significant advances in representation learning have been driven by the development of \gls{cnns} \cite{simonyan2014very, he2016deep, huang2017densely, tan2019efficientnet, liu2022convnet} and \gls{vits} \cite{dosovitskiy2020image, liu2021swin, zhang2023hivit, touvron2021training}. Although \gls{cnns} are effective at capturing local patterns with linear scaling, they struggle to efficiently model global relationships. \gls{vits} address this limitation by introducing self-attention mechanisms, which capture global dependencies by analyzing relationships between all input elements \cite{vaswani2017attention, dosovitskiy2020image}. 
However, the quadratic complexity of \gls{vits} poses challenges for large-scale data processing, as computational costs increase with input size.

\gls{ssms} \cite{gu2021efficiently, fu2022hungry, smith2022simplified} have recently gained attention as a compelling alternative to traditional architectures, offering global receptive fields with linear scalability. Inspired by the achievements of Mamba \cite{gu2023Mamba} in language modeling, recent studies \cite{liu2024vMamba, zhu2024vision, shi2024multi, hatamizadeh2024mambavision} have explored the application of linearly complex global receptive fields to the visual domain. However, unlike textual data, image pixels do not have a sequential dependency. Building on the parallelized selective scan operation in Mamba, known as S6 \cite{gu2023Mamba}, which processes sequential one-dimensional data, VMamba \cite{liu2024vMamba} introduced \gls{ss2d}, a novel four-way scanning mechanism tailored for spatial domain traversal. \gls{vim} \cite{zhu2024vision} further enhanced sequence modeling by employing a bidirectional processing strategy that sums both forward and reverse passes. This process enables the model to capture comprehensive dependencies across an image. Expanding upon these approaches, \gls{msv} \cite{shi2024multi} incorporates a multiscale 2D scanning technique, which processes both original and down-sampled feature maps. This strategy improves capturing long-range dependency while minimizing computational costs. Finally, MambaVision \cite{hatamizadeh2024mambavision} introduces a hybrid approach. Early stages employ \gls{cnns} for fast feature extraction and later stages use Mamba and Transformer blocks to capture both local and global visual features.

\begin{figure*}[!t]
\centering
\includegraphics[trim=0 665 0 0, clip, width=\textwidth]{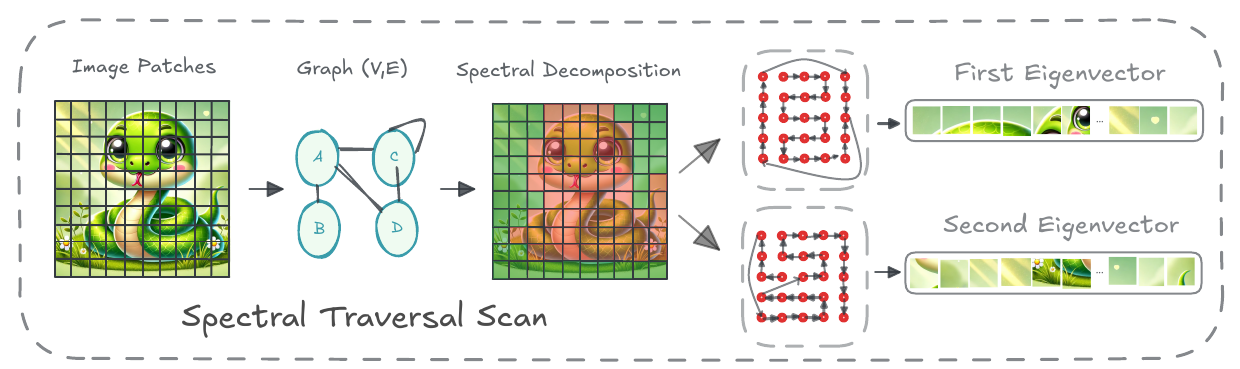}
\caption{The Spectral Traversal Scan (STS) architecture begins by representing image patches as a graph. The Laplacian spectrum of this graph is then generated, and its eigenvectors are computed. The traversal path for the image patches is determined by the order of eigenvectors. The 1st eigenvector traverses the foreground/background regions, the 2nd eigenvector focuses on the next most salient structures, and each subsequent eigenvector progressively captures finer details in the image.}
\label{sts}
\end{figure*}

Vision-based \gls{ssms} are primarily designed to capture relationships between spatially adjacent patches, focusing on spatial proximity or neighborhood-based interactions. However, \gls{ssms} are not inherently equipped to identify relationships between patches that are conceptually related but not spatially adjacent. 
This limitation arises due to the non-causal nature of image data, where pixels lack inherent temporal order or directional relationships.

In addition, current vision-based \gls{ssms} tend to exhibit high sensitivity to spatial information. For instance, VMamba uses predefined traversal routes that cause the network to indiscriminately switch between foreground and background regions within an image. This approach makes the network vulnerable to perturbations and leads to confusion in assessing feature importance. By treating all regions equally, VMamba fails to prioritize the features most relevant to the task, particularly when the foreground contains critical information. This can lead to poor performance when analyzing images with complex compositions. 

Another notable challenge emerges when \gls{ssms} attempt to process images that undergo transformations, such as rotation. Vision-based \gls{ssms} typically process images by (1)~unfolding image patches into sequences, and then (2)~scanning these sequences in specific directions, such as horizontally and vertically. The sequence of image patches depends on the image orientation, meaning that rotating the image alters the order of the sequence. Similarly, the predefined scanning directions, which are fixed to the image original orientation, result in different scan paths when the image is rotated. Consequently, \gls{ss2d} struggle to accommodate transformations such as rotation. Consider Figure \ref{patches} (first row): when the image is rotated by 90 degrees, the predefined and fixed traversal mechanism of VMamba significantly alters the sequence in which patches are processed. 

To address these limitations, we propose \textbf{Spectral VMamba}, which organizes input patches based on the spectral information derived from a graph Laplacian of image patches. By leveraging the eigenvectors from spectral decomposition of an affinity matrix, our approach captures the global structure and relationships within the image. This eigenvector-driven traversal clusters similar patches together, even if they are not spatially adjacent, thereby preventing indiscriminate transitions between foreground and background regions. Moreover, spectral decomposition encodes patch relationships independently of image orientation, enabling the extraction of rotation-invariant features. Hence, our approach defines relationships between patches by their content and mutual associations rather than by their spatial arrangement. Figure~\ref{patches} (second row) demonstrates that rotation does not affect the order in which patches are processed. As we can see, the patches representing the background are processed first, followed by those representing the foreground (the snake in Fig.~\ref{patches}). 

Our main contributions are summarized as follow:
\begin{itemize}
    \item We propose a novel technique named Spectral VMamba based on spectral graph analysis for traversing image patches. By leveraging the eigenstructure of graphs that model the relationships between patches, our approach ensures that spatially-coherent structures in image data are maintained. Our method also improves the performance of \gls{ssms} in capturing and interpreting complex data patterns.
    \item We introduce the \gls{rfn} module, designed to enhance our model robustness to isometric transformations, such as rotations.
    \item Our method demonstrates a notable improvement on the image classification task, outperforming the current state-of-the-art \gls{ssm}-based model in computer vision.
\end{itemize}

\vspace{-10pt}
\section{Related Work}
\label{sec:formatting}

\mypar{State Space Models (SSMs).} 
Transformers have significantly improved the ability to manage long-range dependencies. However, their self-attention mechanism is computationally expensive, especially for high-resolution images, due to the quadratic scaling with input size. To address these challenges, \gls{ssms} have emerged as an effective tool for long sequence modeling \cite{gu2021combining, gu2021efficiently, smith2022simplified, mehta2022long}. In particular, the structured state-space sequence (S4) \cite{gu2021efficiently} model stands out for its efficient handling of long-range dependencies through a diagonal parameterization, which alleviates computational issues seen in earlier models.

Recent advancements in sequence modeling have been driven by innovations in parallel scanning techniques, inspired by models such as S4, S5 \cite{smith2022simplified} and the H3 model \cite{fu2022hungry}. Mamba \cite{gu2023Mamba} improved the S4 model by integrating a data-dependent selection mechanism. This mechanism allows Mamba to achieve linear scalability, ensuring robust performance while accommodating longer sequence lengths without the exponential increase in resource demands. Such improvements are essential for applications requiring real-time or large-scale sequence processing, including \gls{nlp} and time-series forecasting.

\mypar{SSMs in vision.} The success of Mamba models in the \gls{nlp} domain has stimulated significant progress in computer vision, leading to various adaptations tailored for visual data processing. Several studies have extended Mamba architectures to address specific challenges in vision tasks. For instance, VMamba \cite{liu2024vMamba} introduced an \gls{ss2d} module to tackle direction sensitivity between non-causal 2D images and ordered 1D sequences. LocalMamba \cite{huang2024localmamba} focuses on localized feature extraction, utilizing state-space layers to enhance neighborhood interactions and reduce dependence on self-attention, leading to efficient high-resolution image analysis. QuadMamba \cite{xie2024quadmamba} uses a quadrilateral traversal mechanism to manage directional dependencies in 2D images, integrating features from different image segments.

\mypar{Hybrid architectures.} Recent work has also focused on combining SSMs with traditional CNNs or ViTs, leveraging the respective strengths of these models. For example, U-Mamba \cite{ma2024u} merges CNNs and state space models to address long-range dependencies in biomedical image segmentation. MambaVision \cite{hatamizadeh2024mambavision} integrates Mamba architectures with convolutional neural networks to improve performance across various vision tasks by leveraging both sequential modeling and spatial feature extraction. Finally, the TranS4mer model \cite{smith2022simplified} integrates S4 with self-attention mechanisms to achieve state-of-the-art performance in movie scene detection. These advancements collectively demonstrate the versatility and scalability of structured state space models across various applications in the vision domain.

\mypar{Rotation-invariance in vision.} \changes{Numerous approaches have been developed to embed equivariance and invariance directly within neural network architectures. Group-equivariant convolutional networks explicitly encode rotation symmetry into their design, providing intrinsic equivariance to rotations through steerable or group-convolutional filters \cite{cohen2016group, cohen2017steerable, weiler2018learning, e2cnn}. Learned canonicalization methods, such as Spatial Transformer Networks, address invariance by predicting transformations to consistently align input images into a canonical orientation, effectively standardizing pose variability \cite{jaderberg2015stn, kaba2023equivariance, mondal2023equivariant}. Alternatively, symmetrization strategies explicitly aggregate feature responses across multiple rotated inputs or filters, achieving invariance by pooling orientation-specific activations \cite{dieleman2016exploiting, laptev2016tipooling, zhou2017orn, puny2021frame}. Additionally, capsule networks explicitly represent the pose of the object, leveraging pose-aware activations to enhance robustness against rotations \cite{sabour2017dynamic, hinton2018matrix}. Within this landscape, our current work aligns with learned canonicalization approaches by leveraging spectral decomposition to canonicalize patch ordering and achieve patch traversal rotation invariance.}

\vspace{-9pt}
\section{Method}
This section begins with an overview of key concepts in \gls{ssms} and spectral graph analysis in Section \ref{sec:preliminaries}, forming the foundation for our approach. Next, in Section \ref{sec:Spectral VMamba}, we introduce our proposed architecture, Spectral VMamba, along with the \gls{rfn} module designed to achieve rotation invariance (Section \ref{sec:rfn}). Finally, we discuss the robustness of our method to rotational variations in Section \ref{sec:rotation}.

\subsection{Preliminaries}
\label{sec:preliminaries}

\begin{figure*}[!t]
\centering
\includegraphics[trim=0 620 0 0, clip, width=\textwidth]{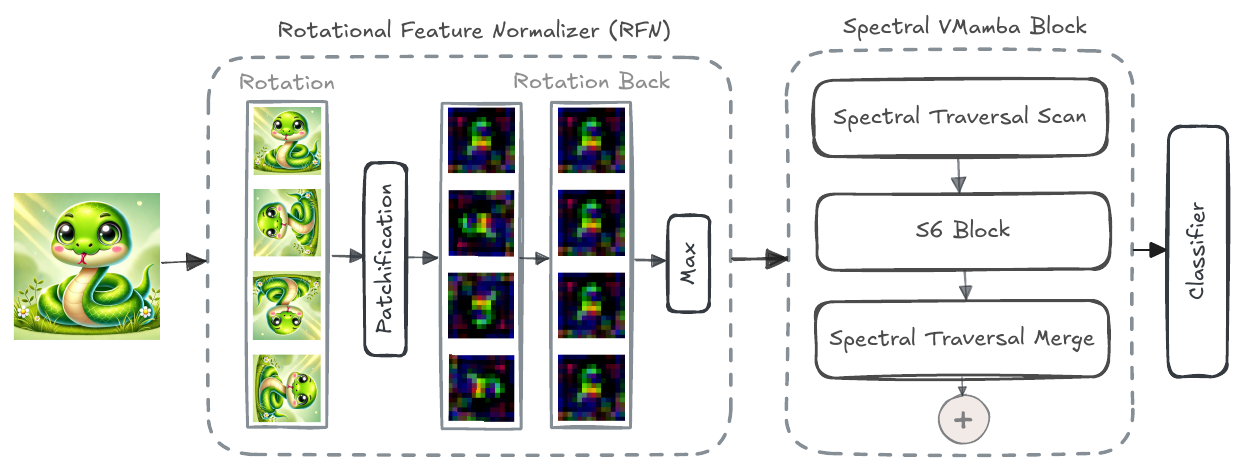}
\caption{An overview of the proposed method. Our network consists of two primary components: the RFN module, which normalizes feature orientation to ensure consistency across different perspectives, and a series of Spectral VMamba blocks that process data in three stages: (1) Spectral Traversal Scan (STS), (2) S6 block (selective scan), and (3) Spectral Traversal Merge (STM). In STS, patches are reordered based on spectral coherence derived from the Laplacian spectrum. Each sequence is then processed in parallel by dedicated S6 blocks, capturing localized patterns, and finally merged in STM to reconstruct the spatial layout, producing the final feature map.}
\label{main}
\end{figure*}

\mypar{SSM formulation.} \gls{ssm}-based models \cite{gu2021efficiently, gu2023Mamba} map a one-dimensional function or an input sequence $x(t) \in \Real$ to an output sequence $y(t) \in \Real$ through a learnable hidden state $\hh(t) \in \Real^N$. This mapping follows a continuous dynamical system formulated as
\begin{equation}
    \frac{d}{dt}\hh(t) = \AAA \hh(t) + \BB x(t), \quad y(t) = \CC \hh(t)
\end{equation}
where $\AAA \in \Real^{N \times N}$, $\BB \in \Real^{N \times 1}$ and $\CC \in \Real^{1 \times N}$ are parameters governing the evolution of system.

To adapt continuous \glspl{ssm} in deep learning frameworks, discretization is required. This can be achieved using the \gls{zoh} rule with a time scale parameter $\Delta \in \Real$. Based on this rule, the discretized versions of $\AAA$ and $\BB$ are defined as
\begin{equation}
\begin{split}
\overline{\AAA} &= \exp(\Delta \AAA), \\
        \overline{\BB} &= (\Delta \AAA)^{-1} \left(\exp(\Delta \AAA) - \II\right) \Delta \BB \approx \Delta \BB
\end{split}        
\end{equation}
and the system becomes
\begin{equation}\label{eq:discrete_ssm}
\hh(t) = \overline{\AAA} \hh(t\!-\!1) + \overline{\BB} x(t), \quad y(t) = \CC \hh(t).
\end{equation}
As $\overline{\AAA}$ and $\overline{\BB}$ are linear operators, the auto-regressive formulation of Eq. (\ref{eq:discrete_ssm}), computing the output at each time step $t$ in a recurrent manner, can be unrolled and expressed as single convolution with kernel $\overline{\KK} \in \Real^{L}$:
\begin{equation}
    \yy = \xx \odot \overline{\KK}, \quad \overline{\KK} = \left( \CC\overline{\BB}, \CC\overline{\AAA\BB}, \ldots, \CC\overline{\AAA}^{L-1}\overline{\BB} \right).
\end{equation}
During training, this convolutional can be employed to compute all output values in parallel.

\mypar{Spectral Graph Analysis} studies the properties of a graph by analyzing the eigen-spectrum of its Laplacian matrix, which can be seen as a discrete version of the continuous Laplace-Beltrami operator obtained by a finite difference method \cite{reuter2006laplace}. Given a real-valued function $f \in C^2$ on a Riemannian manifold $\mathcal{M}$, the Laplace-Beltrami operator $\Delta$ of $f$ is defined as $\Delta f = div(grad\ f)$ where with $grad\ f$ is the gradient of $f$ and $div$ denotes the divergence. 

The Laplacian Eigenvalue Problem corresponds to solving the Helmholtz equation $\Delta f = -\lambda f$, whose solutions (eigenfunctions) are directly related to the vibration of a physical membrane \cite{bayliss1983iterative}. Following Courant's Nodal Line Theorem \citep{courant2008methods}, the Laplacian eigen-spectrum is composed of non-negative eigenvalues $0 \leq \lambda_1 \leq \lambda_2 \leq \ldots$\,, each one repeating according to its multiplicity. Moreover, the eigenfunction corresponding to $\lambda_n$ has at most $n$ poles of vibration, hence smaller eigenvalues and their associated eigenvector encode lower frequency structural information which captures relationships at a more global scale.

The Laplacian matrix of a graph $G = (V,E)$ with nodes $V$ and edges $E$ is defined as $\LL = \DD - \WW $, where $\DD$ is the diagonal degree matrix with entries $ D_{ii} = \sum_j W_{ij}$. A normalized version of the Laplacian is typically employed to address differences in the weight scales $W_{ij}$ and variations in the distribution of node degrees $ D_{ii} $. In this work, we utilize the symmetric normalized Laplacian, defined as
\begin{equation}
\LL_{\text{sym}} \, = \, \II - \DD^{-\frac{1}{2}} \WW \DD^{-\frac{1}{2}},
\end{equation}
which is positive semi-definite and possesses $|V|$ non-negative real-valued eigenvalues $ 0 = \lambda_1 \leq \ldots \leq \lambda_{|V|} $.

A key property of the Laplace eigen-spectrum is that it is invariant to transformations preserving distance (i.e., isometry), which include translations and rotations. In our work, we use this property to design an SSM \changes{that enhances rotational robustness,} where the order in which patches are traversed is determined based on the eigenvectors corresponding to the smallest eigenvalues of the Laplacian matrix.

\subsection{Spectral VMamba}
\label{sec:Spectral VMamba}

Existing vision-based Mamba methods, such as VMamba \cite{liu2024vMamba}, are based on \gls{ss2d}, a four-way scanning mechanism designed for spatial domain traversal. \gls{ss2d} organizes input patches along four predefined scanning routes. However, two key challenges emerge: (1) connections between semantically related but non-neighboring patches are overlooked; and (2) scanning is sensitive to isometric transformations, such as rotation. To address these limitations, we introduce the Spectral VMamba model whose architecture is summarized in Figure \ref{main}. This architecture consists of two main components: a \gls{rfn} module ensuring consistent feature representation across orientations and a sequence of Spectral VMamba blocks forming the network body.

Data forwarding in our Spectral VMamba block comprises three key steps: \gls{sts}, S6 block (selective scan), and \gls{stm}. In the \gls{sts} step, input patches are reordered based on the Laplacian spectrum of their underlying graph structure, ensuring traversal aligns with spectral coherence (see Section \ref{sec:sts}). The selective scan step then processes each reordered patch sequence in parallel using dedicated S6 blocks, effectively capturing localized patterns. Finally, in the \gls{stm} step, the reordered traversal paths are restored to their original spatial format, merging the resulting sequences to generate the final output map. The detailed architecture of \gls{sts} is depicted in Figure \ref{sts}. The sequence of patch processing is determined by the order of eigenvectors derived from the spectral decomposition of the graph Laplacian. It is important to note that \gls{sts} is only performed once for the initial layer. In subsequent layers, we apply a downsampling strategy for generating the traversal path. For a detailed explanation of the downsampling process, please refer to the supplementary material. 

The graph Laplacian eigen-spectrum is isometric invariant, guaranteeing that the order in which patches are traversed is not affected by changes to their position in the image. However, the values in the Laplacian matrix are computed using patch features which are themselves sensitive to rotation of the patch. A simple solution to this problem is to use patches containing a single pixel, however this would be impractical for two reasons. First, individual pixels encode little information (RGB color) and thus using their similarity to construct the Laplacian matrix would not be useful. More importantly, this would severely increase the computational cost of downstream operations such as the spectral decomposition of the Laplacian. To address this issue, we compute patch features using our \gls{rfn} module (Section \ref{sec:rfn}) designed to normalize the feature extraction process by aggregating features from multiple orientations of the input image. This approach allows for \changes{effectively handling of feature changes} during rotations, preserving the desired order and mitigating orientation-dependent discrepancies.


\subsection{Rotational Feature Normalizer (RFN)}
\label{sec:rfn}

An overview of the module is illustrated in Figure \ref{main}. This module rotates the input image by a pre-determined set of angles $\{\theta_i, \ldots, \theta_R\}$ and processes them with a \emph{stem module} which serves as a feature extraction backbone for patchification. This backbone partitions the input image $\II \in \mathbb{R}^{H \times W \times 3}$ into patches $\xx \in \mathbb{R}^{\frac{H}{p} \times \frac{W}{p} \times C}$, where $(H, W)$ denote the dimensions of the input image, $C$ is the number of feature channels, and $p$ represents the patch size. To consolidate features from all rotated versions, each feature map is rotated back to its original orientation. This unrotation ensures that all extracted features align with the original image coordinate system. Finally, an element-wise max operation is applied across the unrotated feature maps to generate a single aggregated feature. The final aggregated feature map can be expressed as:
\begin{equation}
\changes{
{\small
\mathbf{F}_{i,j} = \max_{r \in \{1, \dots, R\}} \left[ \mathcal{R}_{-\theta_r} \big( \text{Patchify} \big( \mathcal{R}_{\theta_r}(\mathbf{I}) \big) \big) \right]_{i,j}
}
}
\end{equation}
where $ \theta_r $ is the angle of the $r$-th rotation and $ \mathcal{R}_{\theta_r} $ is the transformation applying this rotation.

The choice of rotation angles for the \gls{rfn} module represents a trade-off between rotation invariance and computational overhead. Using a larger number of angles ensures a greater coverage in terms of rotation, hence improving the invariance of the model at the cost of increased computations. In our implementation, we used four \emph{canonical} angles, $\{0^\circ, 90^\circ, 180^\circ, 270^\circ\}$, corresponding to quarter turns. By rotating the image at these specific angles, the pixel grid alignment is preserved, eliminating the need for interpolation. This alignment not only simplifies the back-rotation process but also guarantees that the extracted features from all rotated images are comparable.


\subsection{Spectral Traversal Scan} 
\label{sec:sts}

In this section, we describe how the graph Laplacian eigen-spectrum is leveraged to define the travesal order of patches.  

We begin by representing the image as a graph $G = (V, E)$, where each node $v_i \in V$ corresponds to a patch $\xx_i$ of the image, and edges $e_{ij} \in E$ connect pairs of nodes based on their spatial proximity. Following methods for spectral clustering \cite{ng2001spectral} and normalized cuts \cite{shi2000normalized}, we consider a weighted adjacency matrix $\WW$ of the graph, which is constructed using Euclidean distances between patches. Specifically, the weight $W_{ij}$ between the nodes $v_i$ and $v_j$ is given by
\begin{equation}
W_{ij} = \exp\left(-\frac{1}{2\sigma^2}\|\ff_i - \ff_j\|^2\right)
\end{equation}
where $\ff_i$ is the feature vector of patch $\xx_i$ and $\sigma$ is a scaling parameter determined as the average of the Euclidean distances between patches. We set $W_{ij} = 0$ for pairs $(v_i, v_j)$ that are not among the top-$k$ nearest neighbors of each other, resulting in a sparse adjacency matrix.

We compute the first $m$ smallest eigenvectors of the symmetric Laplacian $\LL_{\text{sym}}$ using an iterative method known as the Arnoldi algorithm \citep{golub2013matrix}, which leverages the sparsity of the adjacency matrix to provide a low complexity. Each eigenvector $\uu^{(j)} \in \mathbb{R}^{N_p}$, where $j$ ranges from $1$ to $m$, assigns an eigenfunction value $\uu^{(j)}_i$ to each image patch $i$. In every Mamba block of our model, we perform two traversals per eigenvector: one in the order of increasing $\uu^{(j)}_i$ values and another in decreasing order. At the end of the block, the resulting features from these $2m$ traversals are concatenated together.

Patches $\xx_i$ are ordered based on these eigenvectors, generating multiple traversal sequences:
\changes{
\begin{equation}
{\small
\mathbf{X}^{(j,\mathrm{rev})} = \left\{ \mathbf{x}^{(j,\mathrm{rev})}_{i_1}, \mathbf{x}^{(j,\mathrm{rev})}_{i_2}, \ldots, \mathbf{x}^{(j,\mathrm{rev})}_{i_m} \right\}, \ \text{where}
}
\end{equation}
{\small
\[
i_k^{(j,\mathrm{rev})} = \arg\!\min_{i \in \{1, \ldots, m\}} \left\{ (-1)^{\mathrm{rev}}\!\cdot\! u_i^{(j)} \mid i \notin \{i_1, \ldots, i_{k-1}\} \right\}
\]
}
}
\changes{where $\mathrm{rev}\in\{0,1\}$ indicates if the traversal is reversed ($\mathrm{rev}=1$) or not ($\mathrm{rev}=0$).} This concatenated sequence $\mathbf{X}$ is then fed into the Mamba network for further processing, allowing the model to incorporate rich spectral relationships while preserving global coherence.

\subsection{Spectral Canonicalization} 

Eigenvectors inherently have a sign ambiguity as, if $\uu$ is an eigenvector of a matrix $\AAA$ with eigenvalue $\lambda$, then $-\uu$ is also an eigenvector with the same eigenvalue. To resolve this ambiguity, we standardize the sign of eigenvectors by ensuring that the first non-zero element of each eigenvector is positive. If the first element is negative, we flip the sign of the entire eigenvector. This adjustment does not alter the underlying structure captured by the eigenvectors, as flipping the sign consistently maintains the same eigenvalue $\lambda$ and preserves the relationships between eigenvectors.


\subsection{Rotation Invariance} 
\label{sec:rotation}
Spectral decomposition leverages the eigen-structure of the Laplacian matrix, which remains consistent regardless of rotation. As our traversal strategy is directly guided by the spectral decomposition of input features, the rotation-invariant nature of spectral decomposition extends to our method. This property is demonstrated in the following theorem.

\begin{theorem}
Under the assumption that patch features $\ff$ are invariant to rotation, i.e. $\ff(\II) = \ff(\mathcal{R}_{\theta}\!\left(\II)\right)$ for any $\theta$, our spectral traversal scan is also rotation-invariant.
\end{theorem}
\begin{proof}
As rotations are a type of isometry, we can prove the more general property of isometric invariance. This invariance derives from two properties: 1) the nearest neighbors and values of the weighted adjacency matrix are defined based on distance alone, i.e., $W_{ij} = \exp(-\frac{1}{2\sigma}^2\|\ff_i - \ff_j\|^2)$, thus they do not change if distances remain the same; 2) the eigen-spectrum is invariant under cyclic permutation. This second property can be demonstrated as follows. Let $\PP$ be a permutation matrix and $\uu$ be an eigenvector of Laplacian matrix $\LL$ corresponding to eigenvalue $\lambda$. Using $\uu' = \PP\uu$ (thus $\uu = \PP^T \uu'$) we have that
\begin{equation}
\begin{split}
\LL\uu = \lambda \uu & \, \Rightarrow \, \LL\PP^T\uu' = \lambda \PP^T\uu' \\
& \, \Rightarrow \, \PP\LL\PP^T\uu' = \PP(\lambda \PP^T\uu') \\
& \, \Rightarrow \, \PP\LL\PP^T\uu' = \lambda \uu'
\end{split}
\end{equation}
Hence, $\PP\uu$ is an eigenvector of the permuted Laplacian matrix $\PP\LL\PP^T$. Since we traverse patches by their eigenvector values (from smallest to largest or vice-versa), we will therefore get the same ordering regardless of the permutation.
\end{proof}

\begin{table}[ht!]
\centering
\setlength{\tabcolsep}{5pt}
\renewcommand\arraystretch{1.0}
\caption{Classification performance comparison on \textit{mini}ImageNet. All images are of size $224 \times 224$. T, S, and B denote the tiny, small, and base scales, respectively.}
\label{main_table}
\vspace{-4pt}
\begin{adjustbox}{width=0.9\columnwidth}
\begin{small}
\resizebox{1\linewidth}{!}{
    \begin{tabular}{l|c|c c}
    \Xhline{1.0pt}
    \multirow{2}{*}{Model} & FLOPs & Top-1 & Top-5 \\
    & (G) & (\%) & (\%) \\
    \hline
    \multicolumn{4}{c}{\textbf{Transformer-Based}} \\
    \hline
    DeiT-S~\cite{DeiT2021} & \phz4.6G &  \changes{70.83} & \changes{89.74} \\
    DeiT-B~\cite{DeiT2021}  & 17.5G &  \changes{72.43} & \changes{90.14} \\
    Swin-T~\citep{Swin2021}  & \phz4.5G & 83.25 & 95.54 \\
    Swin-S~\citep{Swin2021}  & \phz8.7G & 84.10 & 95.53 \\
    Swin-B~\citep{Swin2021}  & 15.4G & 82.77 & 95.33 \\
    XCiT-S24~\citep{xcit}  & \phz9.2G & 85.79 & 96.31 \\
    XCiT-M24~\cite{xcit}  & 16.2G & 86.80 & 96.38 \\
    \hline
    \multicolumn{4}{c}{\textbf{SSM-Based}} \\
    \hline
    Vim-T~\cite{zhu2024vision}  & 1.5G  & 67.30 & 87.97 \\
    Vim-S~\cite{zhu2024vision} & 5.1G  & 79.70 & 93.20 \\
    LocalVim-T~\citep{huang2024localmamba}  &  1.5G &  82.12 & 94.60 \\
    LocalVim-S~\citep{huang2024localmamba}  &  4.8G &  81.68 & 93.63 \\
    MSVMamba-N~\citep{shi2024multi}  & 0.9G  &  82.16 & 95.10 \\
    MSVMamba-M~\citep{shi2024multi}  & 1.5G  &  83.72 & 95.81 \\
    MSVMamba-T~\citep{shi2024multi}  & 4.6G  &  86.48 & 96.43 \\
    VMamba-T~\citep{liu2024vMamba}  & 4.9G  & 86.25 & 96.64 \\
    VMamba-S~\citep{liu2024vMamba}  & 8.7G & 86.48 & 96.79 \\
    VMamba-B~\citep{liu2024vMamba}  & 8.7G & 87.17 & 96.96 \\
    
    \hline
    \rowcolor{gray!15}
    Ours-T &  3.9G & 87.86\better{1.61} & 97.25\better{0.61} \\
    \rowcolor{gray!15}
    Ours-S & 6.3G & 88.09\better{1.61} & 97.08\better{0.29} \\
    \rowcolor{gray!15}
    Ours-B & 6.3G & 88.17\better{1.00} &  97.27\better{0.31} \\
    \Xhline{1.0pt}
    \end{tabular}
}
\end{small}
\end{adjustbox}
\vspace{-2pt}
\end{table}

\vspace{-10pt}
\section{Experiments}

We start by comparing our method against recently-proposed ViTs and SSMs on an image classification task.  
We then present an ablation study to thoroughly investigate the impact of various components on the performance of our method.


\subsection{Image Classification}
\label{classification}

We conduct classification experiments on the \textit{mini}ImageNet \cite{vinyals2016matching} dataset. This dataset includes $50,000$ training images and $10,000$ validation images across $100$ categories. Following previous works \cite{Swin2021, liu2024vMamba, zhu2024vision}, we train our models for 300 epochs with a batch size of 128 per GPU. The models are optimized using the AdamW optimizer with a momentum of 0.9. A cosine decay scheduler manages the learning rate, which starts at $5 \times 10^{-4}$, alongside a weight decay of 0.05. Additionally, we apply an exponential moving average (EMA) to stabilize training. For input images of size $224 \times 224$, our data augmentation techniques include color jittering, AutoAugment, random erasing, mixup, and cutmix.


Table \ref{main_table} compares our Spectral VMamba method against \gls{vits} and \gls{ssm}-based models on the \textit{mini}ImageNet dataset. As can be seen, our method outperforms all other approaches in both top-1 and top-5 EMA accuracy. Compared to VMamba, its closest competitor, our method improves top-1 accuracy by 1.61\% for the tiny and small models (T and S) and by 1\% for the base model (B), while requiring less FLOPs (see \ref{ablation} for explanation). Moreover, our Spectral VMamba (B) model yields a 1.37\% higher top-1 accuracy than the strongest ViT-based model, XCiT-M24, and an improvement of 5.4\% over Swin-B, one of the most popular architectures based on ViT.

\subsection{Ablation Study}
\label{ablation}

\mypar{Rotational invariance.} To evaluate the robustness to rotation of our method trained on unrotated images, we evaluated its performance on test images rotated at angles from $0^\circ$ to $360^\circ$ with $15^\circ$ increments. As a reminder, our \gls{rfn} module extracts and aggregates features for four canonical rotations: $\{0^\circ, 90^\circ, 180^\circ, 270^\circ\}$. We therefore expect a greater robustness at those rotation angles, for which our method is fully invariant.



\begin{figure}[t!]
  \centering
   \includegraphics[width=1\linewidth]{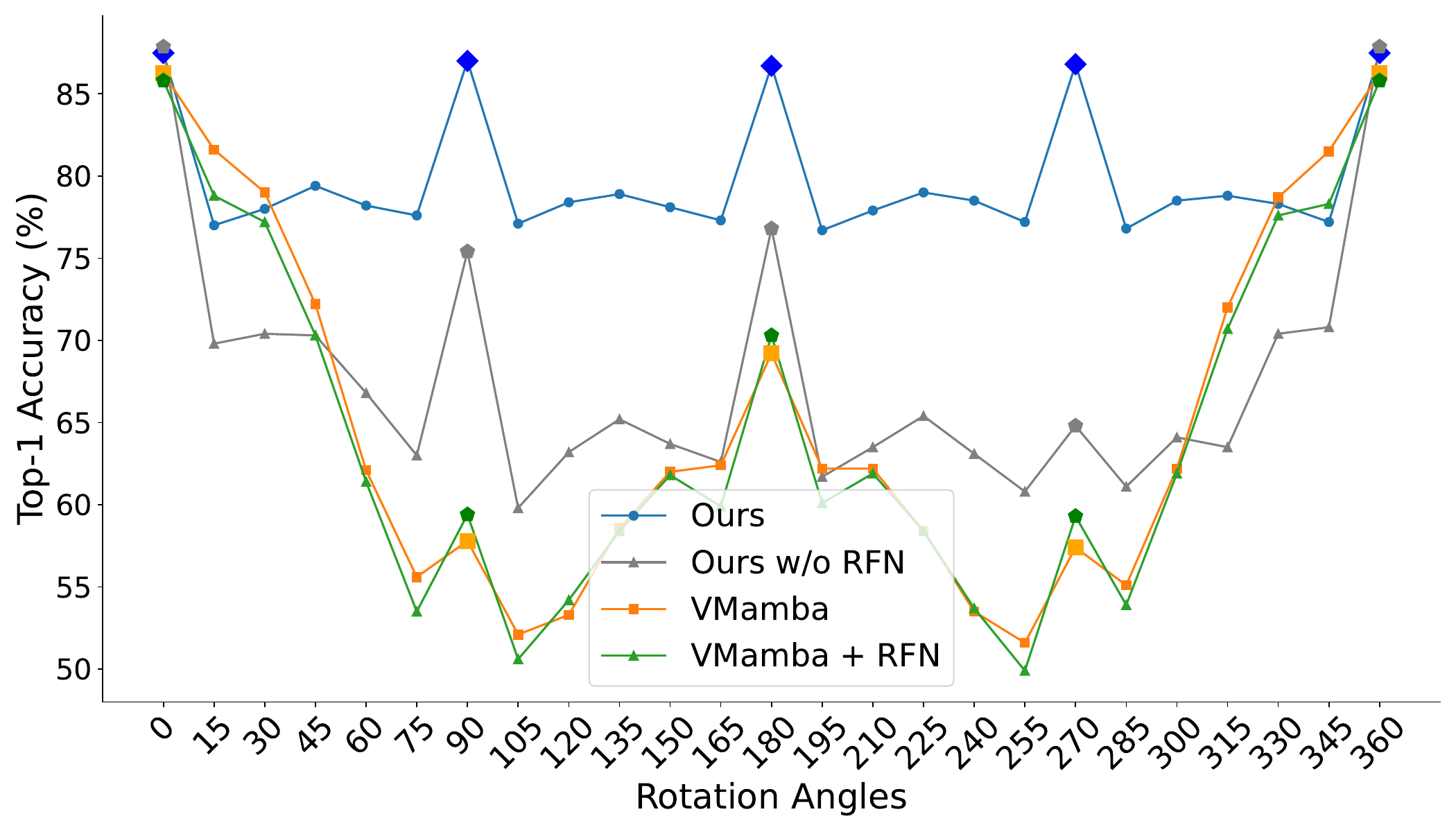}
   \caption{Model performance comparison across various rotation angles. Our method demonstrates consistent accuracy across all rotation angles, while VMamba exhibits significant fluctuations.}
   \label{rotation}
\end{figure}

As shown in Figure~\ref{rotation}, our model with the \gls{rfn} (blue line) maintains a high accuracy near 87\% at those canonical angles, demonstrating its rotational invariance property. In contrast, VMamba (orange line) suffers from a significant drop in performance for images rotated at $90^\circ$ ($\sim$30\% drop), $180^\circ$ ($\sim$20\% drop) and $270^\circ$ ($\sim$30\% drop). Although our method does not guarantee full rotational invariance for non-canonical angles, we observe a remarkably consistent performance across all rotation angles, stabilizing at around 78\% in top-1 accuracy. On the other hand, the performance of VMamba fluctuates drastically, reaching accuracy values near 50\% at certain angles (e.g., $105^\circ$ or $255^\circ$).

\mypar{Impact of \gls{rfn}.} To assess the usefulness of our \gls{rfn} module, we also tested an ablation variant of our method without this module (Ours w/o RFN in Figure \ref{rotation}). As mentioned before, while the Laplacian eigen-spectrum is isometric-invariant, and thus invariant to rotation, the Laplacian matrix is constructed using patch features which are themselves sensitive to rotation. As we expect, removing the \gls{rfn} causes the performance to drop for all angles, this drop growing as we increase the rotation (away from $0^\circ$ or $360^\circ$). Interestingly, our method without the \gls{rfn} module still achieves a higher accuracy than VMamba for angles between $60^\circ$ and $300^\circ$, which represent severe rotations. This could be explained by the small size of patches (16\,$\times$\,16) compared to the image, minimizing the impact of rotations (full invariance would be achieved for 1\,$\times$\,1 patches).  

\mypar{Impact of \gls{sts}.} Since the \gls{sts} is a critical part of our Spectral VMamba method, we cannot remove it. To evaluate its impact, we instead add the \gls{rfn} to VMamba, the resulting approach now only differing from our method by not having a \gls{sts}. As shown in Figure \ref{rotation}, the VMamba + \gls{rfn} model (green line) offers a low robustness to rotation, its accuracy similar to that of VMamba. These results demonstrate that improving rotational invariance at the patch level is not sufficient to achieve a high performance.



\begin{figure}[h!]
  \centering
   \includegraphics[width=.95\linewidth]{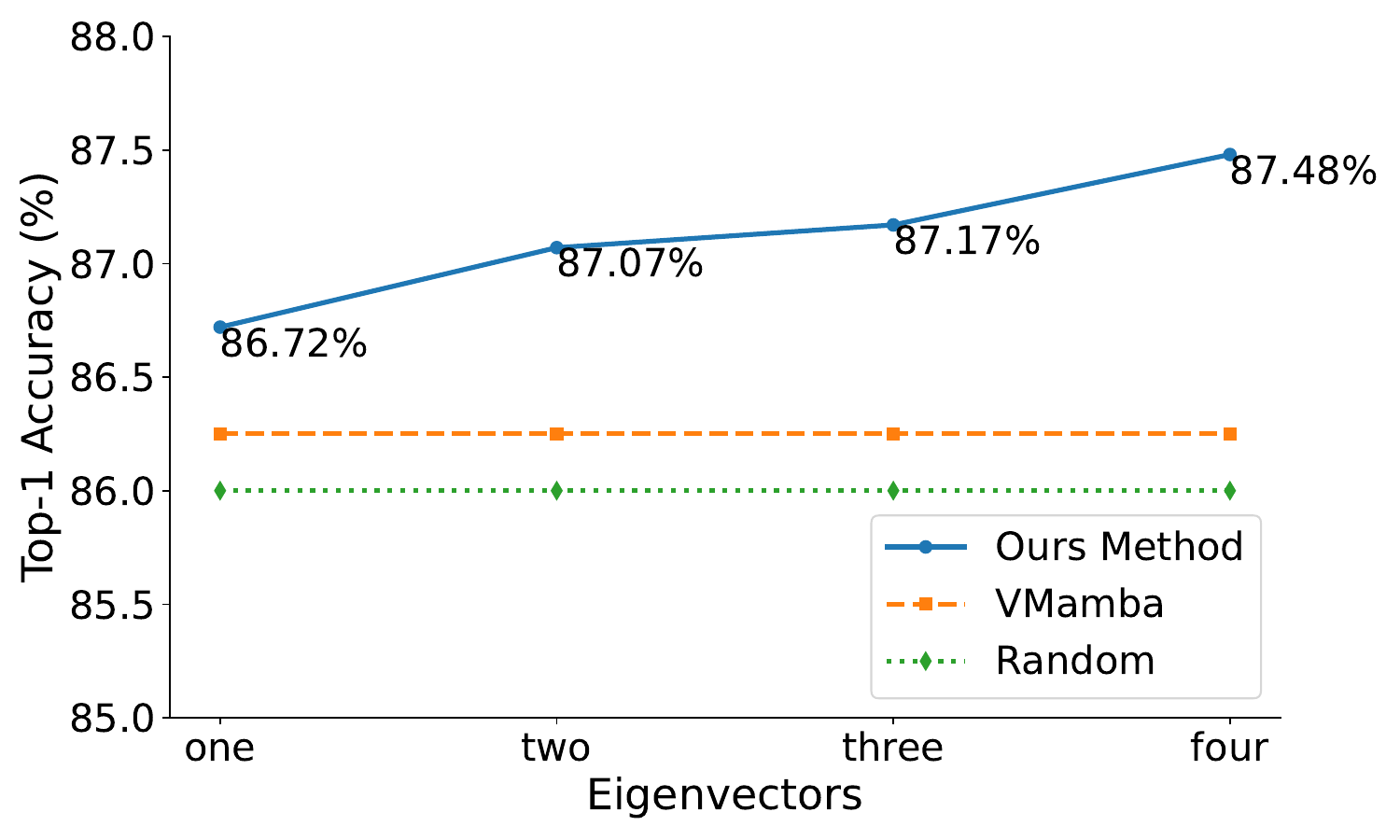}
   \caption{Model performance for different number of eigenvectors. Four eigenvectors exhibit the best performance.}
   \label{eigenvector}
\end{figure}

\mypar{Number of eigenvectors.} In Figure~\ref{eigenvector}, we assess the effect on performance of varying the number of eigenvectors used in the \gls{sts} ($m \in \{1,2,3,4\}$). As can be seen, accuracy improves consistently with more eigenvectors, peaking at 87.48\% for $m=4$. This suggests that eigenvectors offers complementary information and that their combination enhances the recognition of complex global structures. In the same plot, we also report the performance of the standard VMamba model and a variant using a random traversal strategy. As can be observed, the improvement provided by our \gls{sts} over VMamba, for any number of eigenvectors, is greater than the one offered by the raster scan of VMamba over a random traversal.

\mypar{Number of nearest neighbors.} The \gls{knn} algorithm constructs a sparse adjacency matrix by selecting the top $k$ nearest neighbors for each image patch. 
In \cref{tab:accuracy_results}, we report our model's performance across $k$ values from $5$ to $25$. As can be seen, our model is not very sensitive to this parameter and achieves its highest accuracy of 87.48\% for $k\!=\!5$ neighbors. This suggests that only a few neighbors are necessary to capture the global structure of an image. This has a positive impact on the efficiency of our method as the computational complexity of the Laplacian eigen-decomposition step, using the Arnoldi algorithm, is proportional to the number of non-zero entries in the Laplacian matrix (which is directly proportional to $k$).    


\begin{table}[h!]
\centering
\renewcommand{\arraystretch}{1.3} 
\begin{adjustbox}{width=\columnwidth}
\begin{small}
\begin{tabular}{c |c c c c c}
\hline
\textbf{$k$} & 5 & 10 & 15 & 20 & 25 \\
\hline
\textbf{Top-1 Accuracy (\%)} & \textbf{87.48} & 86.98 & 87.14 & 86.91 & 87.15 \\
\hline
\end{tabular}
\end{small}
\end{adjustbox}
\caption{Model performance for different values of $k$ (number of nearest neighbors). The best performance is achieved at $k=5$.}
\label{tab:accuracy_results}
\end{table}


\mypar{Memory Usage and Runtime.} Next, we analyze the computational cost and time efficiency of the \gls{sts}. As mentioned before, the proposed strategy minimizes computations by leveraging the sparsity of the Laplacian matrix and limiting the number of computed eigenvectors. In the following experiments, we use $m=4$ eigenvectors and batch size = 1.

\begin{figure}[h!]
  \centering
   \includegraphics[width=1\linewidth]{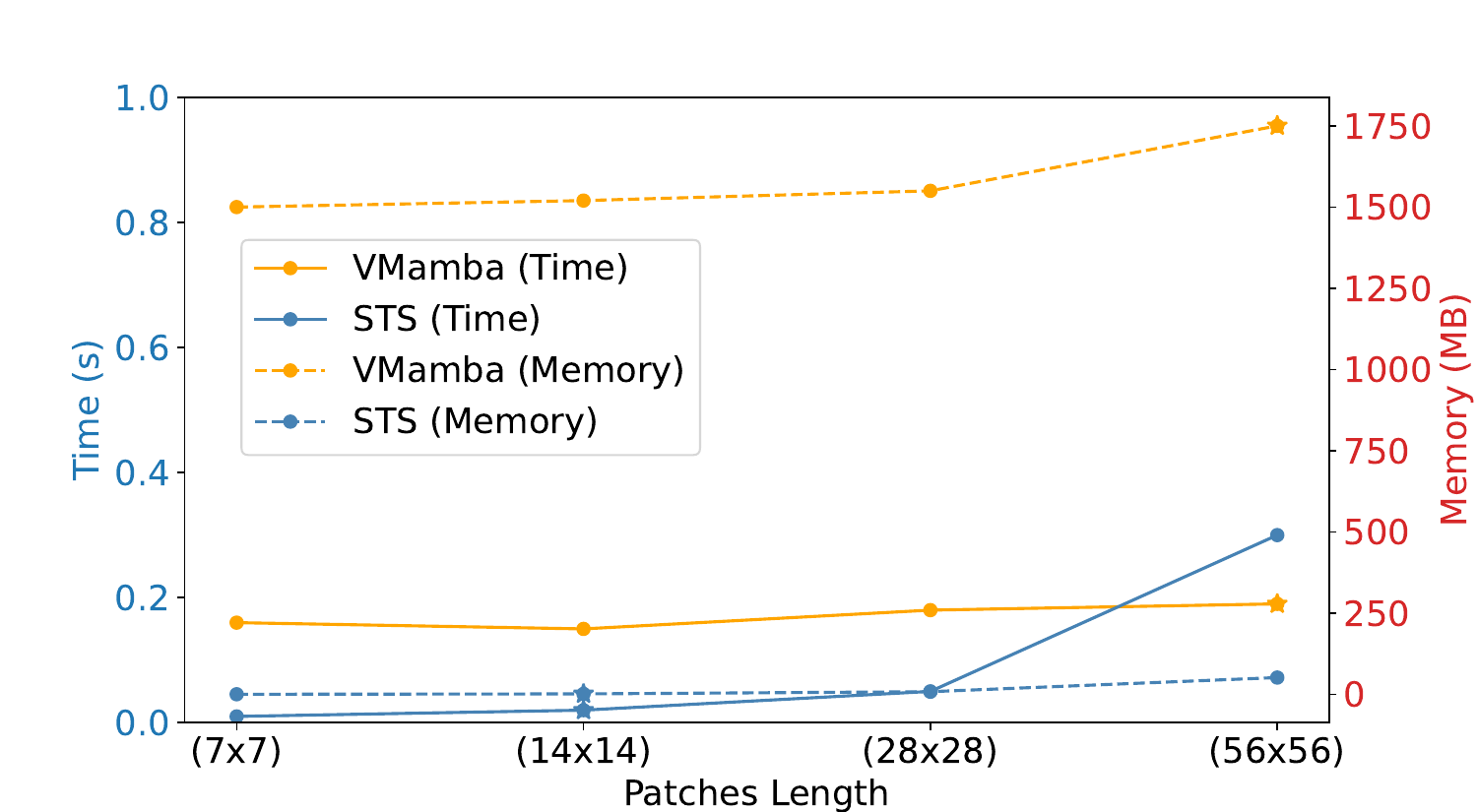}
   \caption{Runtime and memory usage for different patches lengths.}
   \label{memory_time}
\end{figure}

\noindent\textbf{Memory Usage}: As depicted in ~\cref{memory_time}, our \gls{sts} approach (blue line) is highly efficient in terms of memory usage compared to the VMamba (orange line). Even when the number of patches increases, the \gls{sts} based on a sparse eigen-solver, requires only a modest increase in memory compared to the VMamba. The stars along each line indicate the token lengths used in our method, highlighting the specific points where processing occurs across different sequence lengths.

\noindent\textbf{Runtime}: Our \gls{sts} strategy, which can run on a CPU, also shows a favorable runtime scaling with increased patch length (\cref{memory_time}). Specifically, for a patch length of 196 (marked by a star), which is the  setting used in our main experiments, the increase in runtime is minimal when compared to the VMamba backbone, highlighting the efficiency of our approach. 

\noindent\textbf{FLOPs}: The amount of FLOPs for our traversing strategy include the computations required for constructing the adjacency matrix, calculating the Laplacian, and obtaining its first eigenvectors. This calculation results in approximately 2 MB of FLOPs, which is significantly lower than the 3.9 GB FLOPs needed by VMamba's backbone. Interestingly, as reported in \cref{main_table}, the total number of FLOPs performed by our Spectral VMamba method is \emph{less} than that of VMamba when using the same model size. This is because, after processing patches, the input to our Spectral VMamba has a size of $14\!\times\!14$ tokens, compared to $56\!\times\!56$ tokens for VMamba.

\section{Conclusion}

In this paper, we introduce Spectral VMamba, a novel approach to visual state space models that achieves \changes{patch traversal} rotation invariance. Our method leverages spectral decomposition from the graph Laplacian of image patches to define traversal paths for patch processing. In addition, we develop the \gls{rfn} module to ensure consistent feature representation in different orientations. This module serves as a foundation for our \gls{sts} module, enabling patch features that are robust to rotational variations. By incorporating the \gls{rfn} and \gls{sts} modules, Spectral VMamba achieves a robust transformation-invariant feature extraction pipeline that maintains high accuracy across a variety of orientations. Our experiments demonstrate that Spectral VMamba consistently outperform popular models including Swin transformer, VMamba,  MSVmamba, and LocalVim in image classification tasks. Our method does not yet address generalization to the medical domain. In future work, we aim to handle disconnected graph components, common in medical images with distinct anatomical structures or pathologies. Adapting spectral traversal for these regions will enhance pattern capture and robustness across imaging modalities.



\section*{Acknowledgement}

We appreciate the computational resources and support provided by Compute Canada and the Digital Research Alliance of Canada.

{
    \small
    \bibliographystyle{ieeenat_fullname}
    \bibliography{main}
}


\clearpage
\setcounter{page}{1}
\setcounter{section}{0}
\renewcommand{\thesection}{\Alph{section}}
\renewcommand{\thesubsection}{\thesection.\arabic{subsection}}
\maketitlesupplementary

\section{Implementation Details}

In this section, we give the pseudo-code for the \gls{rfn} and \gls{sts} modules.
This pseudo-code provides a concise summary of the key steps involved in our approach, offering a high-level abstraction of the implementation. It is designed to complement the detailed explanations in the main paper and can serve as a reference for reproducing our results.

We begin with Algorithm \ref{rfn_pseudo}, which outlines the steps for implementing the \gls{rfn} module. As discussed in the paper, this module ensures a consistent representation of image features across different orientations. This module comprises four key steps: rotation, patchification, back-rotation, and a max operation to aggregate features across all orientations.

\begin{algorithm}
\caption{Rotational Feature Normalizer Module}
\label{rfn_pseudo}
\begin{algorithmic}[1]
\Require Input image $\mathcal{I} \in \mathbb{R}^{H \times W \times 3}$; rotation angles: $\{\theta_r \mid r = 1, \dots, R\}$; stem module: $\text{Stem}$; patch size: $p$
\Ensure Aggregated feature map $\mathcal{F}$.

\State \textbf{Initialize:} $\{\theta_r\}$ and $\{-\theta_r\}$.

\For{$r \in \{1, \dots, R\}$}
    \State $\mathcal{I}_r \gets \mathcal{R}_{\theta_r}(\mathcal{I})$ \Comment{Rotate image by $\theta_r$}
    \State $\mathcal{F}_r \gets \text{Stem}(\mathcal{I}_r)$ \Comment{Extract features}
    \State $\mathcal{F}_r^{\text{unrotated}} \gets \mathcal{R}_{-\theta_r}(\mathcal{F}_r)$ \Comment{Back-rotate feature map}
    \State Append $\mathcal{F}_r^{\text{unrotated}}$ to $\mathcal{F}_r$.
\EndFor
\State $\mathcal{F} \gets \max_{r \in \{1, \dots, R\}} \mathcal{F}_r$ \Comment{Max operation}

\State \textbf{Output:} Aggregated feature map $\mathcal{F}$.
\end{algorithmic}
\end{algorithm}

Algorithm \ref{sts_pseudo} represents the \gls{sts} strategy for determining the traversal order of image patches based on spectral decomposition. The process begins by constructing the adjacency matrix $\WW$ using the $k$-Nearest Neighbors (KNN) algorithm, based on the Euclidean distances between image patches. This matrix captures the relationships and similarities between the patches. Subsequently, the degree matrix $\DD$ is computed, where each diagonal entry $D_{ii}$ represents the sum of the weights of edges connected to node $i$. Using $\WW$ and $\DD$, the symmetric normalized Laplacian matrix $\LL_{\text{sym}}$ is calculated. Spectral decomposition is then applied to $\LL_{\text{sym}}$, yielding the eigenvalues $U$ and eigenvectors $V$. Finally, the patches are reordered based on the spectral information, resulting in the ordered sequence of patches $P$.

\begin{algorithm}
\caption{Spectral Traversal Scan Module}
\label{sts_pseudo}
\begin{algorithmic}[1]
\Require patch feature $\ff$, number of neighbors $k$, and eigenvectors $m$
\Ensure Traversal sequence $\mathcal{P}$

\vspace{3pt}
\State \textbf{Step 1}: \emph{Compute Weighted Adjacency Matrix}
\For{each pair of patches $(\xx_i, \xx_j)$}
    \If{$i \in knn_j$ OR $j \in knn_i$}
        \State $W_{ij} = \exp\left(-\frac{\|\ff_i - \ff_j\|^2}{2\sigma^2}\right)$
    \Else
        \State $W_{ij} = 0$
    \EndIf
\EndFor

\vspace{3pt}
\State \textbf{Step 2}: \emph{Compute Normalized Laplacian}

\vspace{3pt}
\State $\LL_{\text{sym}} \, = \, \II - \DD^{-\frac{1}{2}}\WW\DD^{-\frac{1}{2}}$

\vspace{3pt}
\State \textbf{Step 3}: \emph{Compute $m$ Smallest Eigenvectors of $\LL_{\text{sym}}$}

\vspace{3pt}
\State $[U, V] = \text{eigenSolver}(L_\text{sym}, m)$

\vspace{3pt}
\State \textbf{Step 4}: \emph{Building Traversal Sequences}

\vspace{3pt}
\State Sort $U$ from smallest to largest.

\vspace{3pt}
\State Select the corresponding $m$ eigenvalues from $V$.

\vspace{3pt}
\For{$j = 1$ to $m$}
    \State $\text{P}_1^{j} = \text{sorting $f$ using increasing order of} \ \vv^{(j)}$
    \State $\text{P}_2^{j} = \text{sorting $f$ using decreasing order of} \ \vv^{(j)}$
    \State $\mathcal{P} = \left[ P_1^{(j)}, \; P_2^{(j)} \right]^{j=1, \ldots, m}$
\EndFor

\State \textbf{Output:} Traversal sequence $\mathcal{P}$.
\end{algorithmic}
\end{algorithm}


\section{Downsampling Strategy}

Similar to VMamba, the architecture of our network consists of four layers. Each layer contains a different number of Spectral VMamba blocks. For example, in the tiny scale configuration, we use the setup [2, 2, 5, 2], indicating that the first layer has two Spectral VMamba blocks, the second layer also has two blocks, the third layer has five blocks, and the final layer contains two blocks. As previously mentioned, the \gls{sts} module is applied only once, in the first Spectral VMamba block of the first layer. Additionally, downsampling occurs at the end of each layer.

The eigenvectors are initially computed in the \gls{sts} using the original $14 \times 14$ spatial features. However, after downsampling, the eigenvectors derived from the $14 \times 14$ patches no longer align with the downsampled patches in subsequent layers, which requires careful handling to maintain consistency. To resolve this alignment issue, we save the indices used during the max pooling operation, which is responsible for downsampling the spatial features. By leveraging the saved indices, we extract the corresponding eigenvectors, generating a new set that matches the shape of the downsampled patches. This alignment ensures that the eigenvectors and patches remain correctly paired, enabling us to order the downsampled patches using eigenvectors of compatible dimensions. We repeat this process at each subsequent layer to maintain proper alignment throughout the network.


\begin{figure}[ht!]
\centering
\includegraphics[trim=0 560 0 0, clip, width=.47\textwidth]{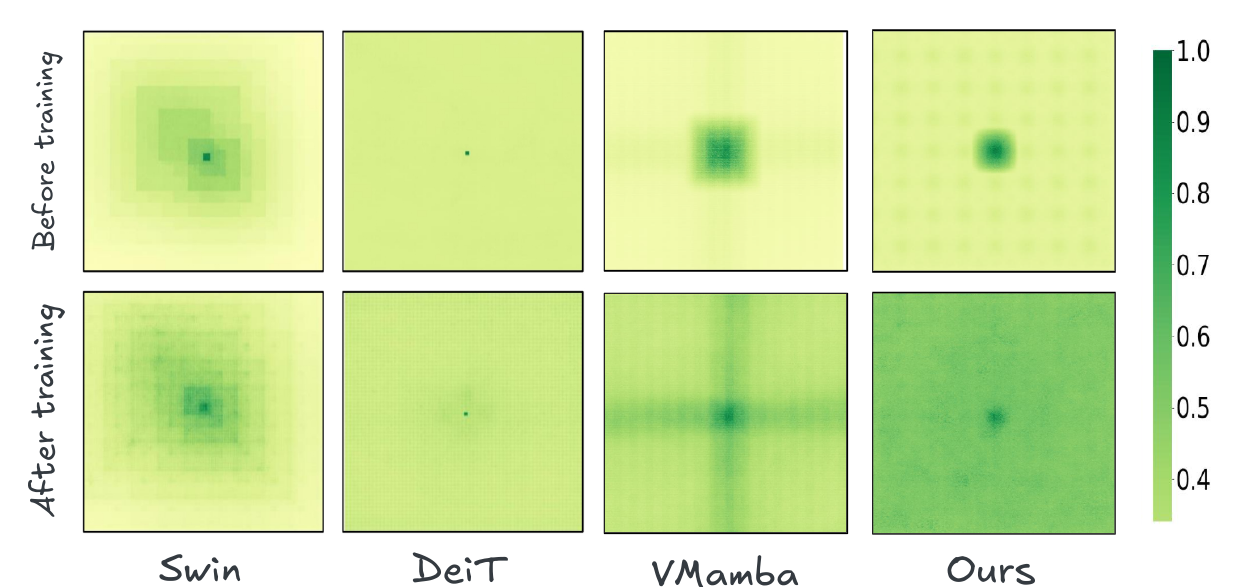}
\caption{Comparison of Effective Receptive Fields (ERF) Before and After Training.}
\label{erf}
\end{figure}

\section{Visualization of Activation Maps}

\gls{erf} in \gls{cnns} refer to the region of the input image that significantly influences the activation of a particular neuron in a deeper layer of the network. Unlike the theoretical receptive field, which considers the full extent of influence regardless of intensity, the \gls{erf} focuses on the practical impact, often showing that only a central portion of the theoretical receptive field has substantial influence. Analyzing \gls{erf}s helps in refining network designs to ensure that neurons capture relevant information efficiently, enhancing performance in tasks such as object detection and image segmentation.

We conducted experiments to compare our model with VMamba, focusing on the \gls{erf} of the central pixel before and after training. Our results in Figure~\ref{erf} indicate that our model exhibits more global \gls{erf}s compared to VMamba. Specifically, after training, the areas colored dark green, which represent regions of high influence, are more extensive in our method than in VMamba. This suggests that our model is better at capturing broader contextual information from the input image. The increased dark green regions in our model demonstrate its enhanced capability to integrate information over larger portions of the input, potentially leading to improved performance in tasks requiring a comprehensive understanding of the image content.

\begin{figure}[ht!]
  \centering
   \includegraphics[width=\linewidth]{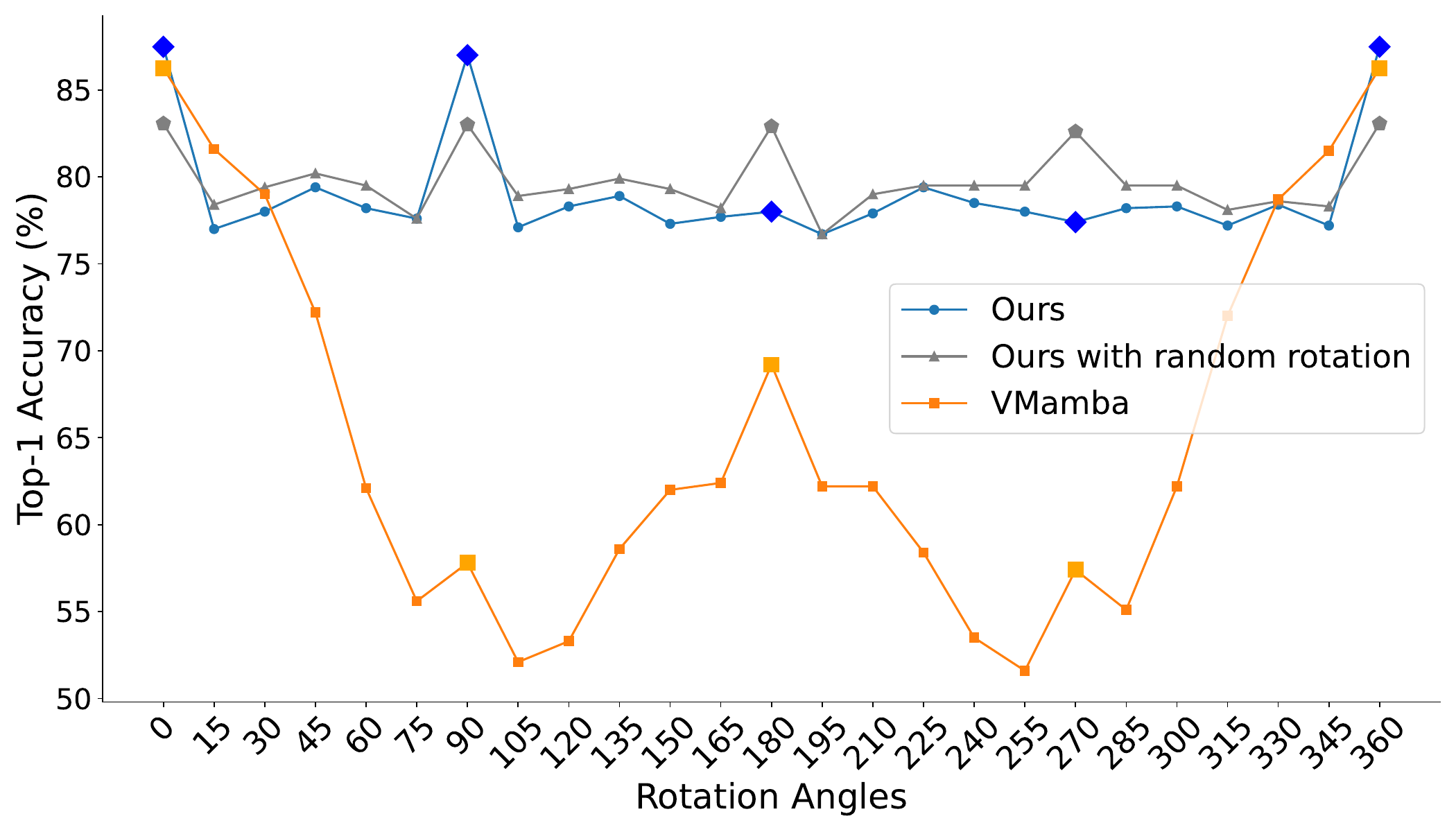}
   \caption{Effect of Random Rotation in RFN Module.}
   \label{random}
\end{figure}

\section{Training Dynamics}

\begin{figure*}[ht!]
  \centering
   \includegraphics[width=\linewidth]{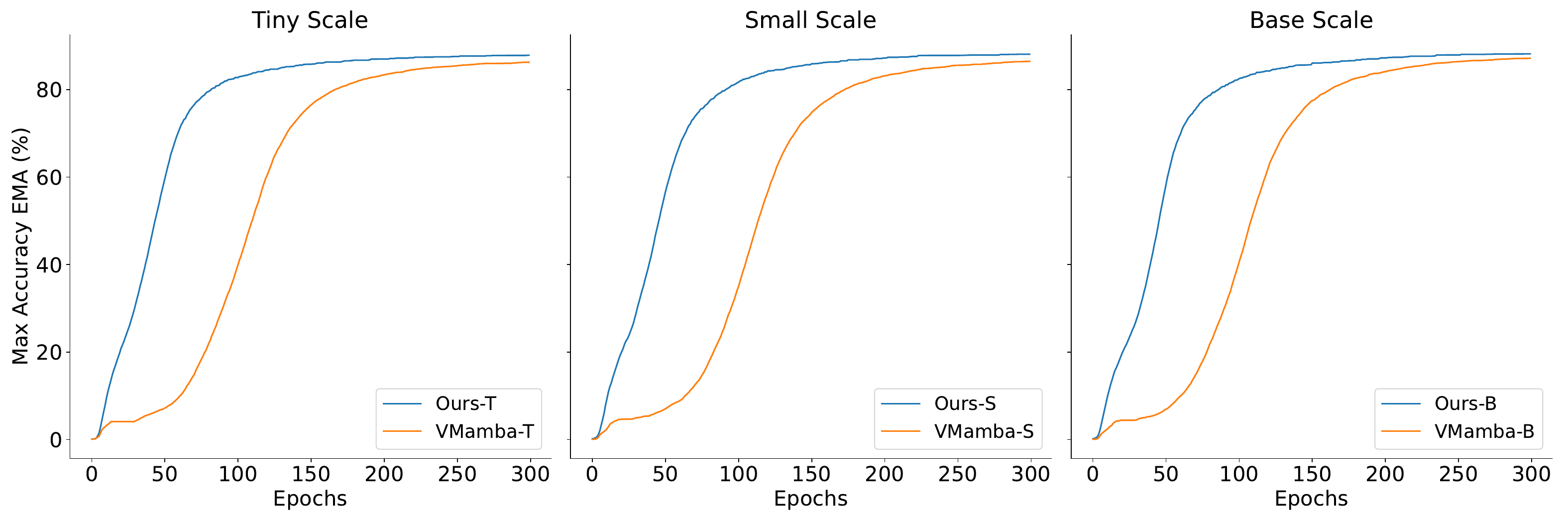}
   \caption{Training Speed Comparison.}
   \label{speed}
\end{figure*}

Figure \ref{speed} illustrates the comparison of training dynamics, measured using the maximum accuracy with an Exponential Moving Average, across three model scales: Tiny, Small, and Base. Each plot represents the accuracy progression over $300$ epochs for our proposed method and the VMamba model. The plots represents our method demonstrates faster convergence across all scales, achieving high accuracy earlier in training compared to VMamba. The faster convergence not only highlights the efficiency of our approach but also reduces training time, making it highly suitable for practical applications where computational resources or time are constrained.




\section{RFN with Random Rotations}

For the \gls{rfn} module, we utilized four \emph{canonical} angles: ${0^\circ, 90^\circ, 180^\circ, 270^\circ}$, corresponding to quarter turns. Additionally, we tested the method with four random rotations selected at each iteration from the ranges [0, 90], [91, 180], [181, 270], and [271, 360]. Unlike quarter turns, most random rotations necessitate interpolation. The results, presented in Figure \ref{random}, demonstrate that even with random rotations, the model can achieve a high level of invariance. 
However, a slight accuracy drop was observed when using random rotations (83.06\%) compared to the canonical quarter turns (87.48\%), which can be due to the loss of details resulting from interpolation.


\section{Downstream tasks}

To demonstrate the versatility of our pre-trained model and its applicability to tasks beyond classification, we extended its use to semantic segmentation. Specifically, we fine-tuned the model—originally pre-trained on the \textit{mini}ImageNet dataset—to perform segmentation tasks on ADE20K dataset \cite{zhou2019semantic}. ADE20K includes 150 fine-grained semantic categories and comprises 20,000 training images, 2,000 validation images, and 3,000 test images. For optimization, we use AdamW with a weight decay of 0.01 and a total batch size of 2 per GPU. The training schedule features an initial learning rate of 6\,$\times$\,10$^{-5}$, linear decay, a 1500-iteration linear warmup, and a total of 160,000 iterations. We apply standard data augmentations such as random horizontal flipping, random scaling within a ratio range of 0.5 to 2.0, and random photometric distortion.

\changes{
Table \ref{segmentation} shows that our method outperforms VMamba by +2.81 (tiny), +1.68 (small), and +1.49 (base) on single-scale (SS) testing, and by +3.74 (tiny), +2.21 (small), and +0.58 (base) on multi-scale testing. The backbone used for the segmentation task was initialized from a classification checkpoint which was trained on the mini-ImageNet dataset. This explains the relatively lower range of segmentation performance compared to other papers, which often use classification models pre-trained on ImageNet-1k. }

\begin{table}[ht]
\centering
\begin{tabular}{@{}lcc@{}}
\toprule
\textbf{Method}    & \textbf{mIoU (SS)}           & \textbf{mIoU (MS)}\\ \midrule
VMamba-T           & 22.77                              & 23.98\\ 
VMamba-S           & 25.84                              & 27.13\\ 
VMamba-B           & 26.32                              & 28.41\\ \midrule
Ours-T             & 25.58 \textbf{\better{2.81}}       & 27.72 \textbf{\better{3.74}}\\ 
Ours-S             & 27.52 \textbf{\better{1.68}}       & 29.34 \textbf{\better{2.21}}\\ 
Ours-B             & 27.81 \textbf{\better{1.49}}       & 28.99 \textbf{\better{0.58}}\\ 
\bottomrule
\end{tabular}
\caption{Results of semantic segmentation on ADE20K. SS and MS denote single-scale and multi-scale testing, respectively.}
\label{segmentation}

\end{table}

\end{document}